\documentclass[10pt,twocolumn,letterpaper]{article}

\usepackage{cvpr}
\usepackage{times}
\usepackage{epsfig}
\usepackage{graphicx}
\usepackage{amsmath}
\usepackage{amssymb}
\usepackage{bm}
\usepackage{caption}
\usepackage{enumerate}
\usepackage[ruled]{algorithm2e}
\usepackage[table,xcdraw]{xcolor}
\usepackage{color}
\usepackage{enumitem}
\usepackage[numbers]{natbib}
\usepackage{float}
\usepackage{subfigure}
\usepackage{threeparttable}
\usepackage{multirow}
\usepackage{booktabs}
\usepackage[table]{xcolor}
\usepackage{arydshln}
\usepackage{threeparttable}
\usepackage{colortbl}
\usepackage{enumitem}
\usepackage{array}
\usepackage[table]{xcolor}
\usepackage{colortbl}
\usepackage{url}

\def\ourmodel{UMA}

\usepackage[pagebackref=true,breaklinks=true,letterpaper=true,colorlinks,bookmarks=false]{hyperref}

\cvprfinalcopy 

\def\cvprPaperID{4820} 

\ifcvprfinal\fi
\begin{document}

\title{A Unified Object Motion and Affinity Model for Online Multi-Object Tracking}
\author{Junbo Yin\textsuperscript{1}, Wenguan Wang$^{2}\thanks{Corresponding author: \textit{Wenguan Wang}.}$~\!, Qinghao Meng\textsuperscript{1}, Ruigang Yang\textsuperscript{3,4,6}, Jianbing Shen\textsuperscript{5,1}\\
\small{\quad$^1$}\small Beijing Lab of Intelligent Information Technology,
School of Computer Science, Beijing Institute of Technology, China \\
\!\!\small{$^2$}\small ETH Zurich, Switzerland~~
\small{$^3$} \small Baidu Research~~ \small{$^4$} \small National Engineering Laboratory of Deep Learning Technology and Application, China \quad \\
\small{$^5$} \small  Inception Institute of Artificial Intelligence, UAE \quad
\small{$^6$} \small University of Kentucky, Kentucky, USA\\
{\tt\small yinjunbo@bit.edu.cn \quad wenguanwang.ai@gmail.com}\\
	{\tt\small \url{https://github.com/yinjunbo/UMA-MOT}}
}

\maketitle
\thispagestyle{empty}

\begin{abstract}
Current popular online multi-object tracking (MOT) solutions apply single object trackers (SOTs) to capture object motions, while often requiring an extra affinity network to associate objects, especially for the occluded ones. This brings extra computational overhead due to repetitive feature extraction for SOT and affinity computation. Meanwhile, the model size of the sophisticated affinity network is usually non-trivial. In this paper, we propose a novel MOT framework that unifies object motion and affinity model into a single network, named \ourmodel, in order to learn a compact feature that is discriminative for both object motion and affinity measure. In particular, \ourmodel~integrates single object tracking and metric learning into a unified triplet network by means of multi-task learning. Such design brings advantages of improved computation efficiency, low memory requirement and simplified training procedure. In addition, we equip our model with a task-specific attention module, which is used to boost task-aware feature learning. The proposed \ourmodel~can be easily trained end-to-end, and is elegant -- requiring only one training stage. Experimental results show that it achieves promising performance on several MOT Challenge benchmarks.
\end{abstract}

\vspace{-8pt}
\section{Introduction}\label{sec:intro}
Online multi-object tracking (MOT) aims to accurately locate trajectory of each target while maintaining their identities with information accumulated up to the current frame. In the last decades, MOT has attracted increasing attentions, as it benefits a wide range of applications, such as video surveillance analyses and autonomous driving~\cite{song2019apollocar3d,wang2018semi,wang2019learning}.

Current MOT solutions typically involve an object motion model and an affinity model. The former one leverages temporal information for object instance localization and tracklet generation, while the latter deals with distractors (\eg, targets with similar appearance) or occlusions by measuring object similarity in data association. Specifically, some online MOT algorithms are based on a tracking-by-detection paradigm~\cite{hong2016online,milan2016online,tang2017multiple,bae2014robust,kim2018multi}, \ie, associating detections across {{frames by computing pairwise affinity.}} Thus they mainly focus on the design of the affinity model. However, as temporal cues are not explored in the object detection phase, the quality of detections {{is}} often limited, further decreasing the MOT performance. MOT scenarios, \eg, the video sequences in {MOT Challenge}~\cite{leal2015motchallenge,milan2016mot16}, often yield crowded people with rare poses or various sizes. In such cases, even the leading detectors~\cite{ren2015faster} may produce many False Positive (FP) and False Negative (FN) results, causing adverse impact on the subsequent data association stage.


This calls for better leveraging motion cues in MOT. Thus another trend is to apply single object trackers (SOTs) in online MOT~\cite{chu2017online,wojke2017simple}. They take advantage of SOTs to address the value of temporal information and recover the missing candidate detections. {Such paradigm} yields more natural tracklets and usually leads to better tracking results according to the FN metric. However, crowded distractors and their frequent interactions often lead to occlusion situations, which are quite challenging for these solutions. To tackle this issue, follow-up methods~\cite{sadeghian2017tracking,chu2017online,zhu2018online,chu2019online,chu2019famnet} integrate SOT based motion model with affinity estimation. In particular, they first recognize the state of the targets according to the confidence of SOTs, then update the tracked targets and maintain the identities of the occluded targets through affinity measure for tracklet-detection pairs in data association phase. Though inspired, they still suffer from several limitations. First, features used for SOTs and affinity measure are extracted from two separate models, which incurs expensive computational overhead. Second, since they do not make use of SOT features in affinity computation, they have to train an extra affinity network (\eg, ResNet50 in~\cite{zhu2018online} and ResNet101 in~\cite{chu2019famnet}) to remedy this. This further increases their memory demand, which critically limits their applicability in source-constrained environments. Third, the independent feature extraction of SOTs and affinity models, and the complicated affinity network design, together make the training procedures sophisticated, which often require multiple alternations or cascaded-training strategy. Moreover, they do not explore the relation of the SOTs and the affinity model, \ie, the affinity model could help the SOTs to access the identity information and thus learn more discriminative features to better handle occlusions.

To alleviate the above issues, we propose a multi-task learning based online MOT model,~\ourmodel, which end-to-end integrates the SOT based motion model and affinity network into a unified framework. The learnt features are promoted to capture more identity-discriminative information, thus simplifying the training and testing {{procedures.}} In particular, it unifies a Siamese SOT and a ranking network into a triplet architecture. Two branches of the triplet network, \eg, the positive and anchor branches, count for the SOT based motion prediction task, while all the three branches address the target identity-aware ranking task by metric learning. This provides several benefits. First, the metric learning within the ranking task assigns the learnt features identity-discriminative ability, facilitating the SOT model to better locate the targets and handle occlusions. Second, this enables feature sharing between the SOT based tracklet generation stage and the affinity-dependent data association stage, eliminating the requirement of designing an additional affinity network and improving the computation efficiency. Third, it provides a more straightforward, one-step training protocol instead of previous sophisticated, multi-alternation or cascaded training strategy. Furthermore, a task-specific attention (TSA) module is equipped with our~\ourmodel~model, to address the specific nature of the multiple tasks and boost more task-specific feature learning. It performs context exploitation self-adaptively on the shared features extracted by the multi-task network and is lightweight with budgeted computational cost, {{meanwhile producing}} better performance. To summarize, we propose a triplet network,~\ourmodel, which unifies the object motion prediction and affinity measure tasks in online MOT. \ourmodel~addresses SOT-applicable as well as association-discriminative feature learning with an attentive multi-task learning mechanism. This presents an elegant, effective yet efficient MOT model with lower memory requirement and a simple, end-to-end training protocol. Further with an elaborately designed online tracking pipeline, our lightweight model reaches state-of-the-art performance against most online and even offline algorithms on several MOT Challenge benchmarks.



\section{Related Work}
\noindent\textbf{MOT.}~Existing MOT approaches can be categorized into offline and online modes. \textit{Offline} methods~\cite{pirsiavash2011globally,dehghan2015target,tang2016multi,wang2016joint,tang2017multiple} can leverage both past and future frames for batch processing. They typically consider MOT as a global optimization problem in various forms such as multi-cut~\cite{tang2016multi,tang2017multiple}, k-partite graph~\cite{Zamir2012GMCP,dehghan2015gmmcp} and network flow~\cite{zhang2008global,dehghan2015target}. Though favored in handling ambiguous tracking results, they are not suitable for causal applications such as autonomous driving.

\textit{Online} MOT methods can only access the information available up to the current frame, thus easily suffering from target occlusions or noisy detections. The majority of previous approaches~\cite{bae2014robust,bergmann2019tracking, hong2016online,milan2016online,xu2019spatial} adopt a tracking-by-detection pipeline, whose performance is largely limited by the detection results. Some others~\cite{zhu2018online,sadeghian2017tracking,chu2017online,chu2019online,chu2019famnet} instead apply SOTs~\cite{henriques2014high,bertinetto2016fully,liu2016structural,dong2016occlusion,dong2019dynamical} to carry out online MOT and generally gain better results. 

\noindent\textbf{Object Motion Models in Online MOT.} Basically, object motion model is helpful in dealing with the noisy detections. For instance, Xiang \textit{et al.}~\cite{xiang2015learning} employ an optical flow-based SOT, TLD~\cite{kalal2012tracking}, to track individual target. Sadeghian \textit{et al.}~\cite{sadeghian2017tracking} further extend this pipeline with a multi-LSTM network for exploiting different long-term cues. After that, Zhu \textit{et al.}~\cite{zhu2018online} equip their framework with a more advanced tracker: ECO~\cite{danelljan2017eco}, and design an attention-based network to handle occlusions. Their promising results demonstrate the advantages of applying SOTs as motion models. However, all these approaches require an additional affinity model to address occlusions, and often learn features for the SOTs and affinity models independently, leading to increased computation cost, non-trivial memory demand and sophisticated training protocols. Though~\cite{chu2017online} uses a shared backbone to extract features for all the targets, multiple online updating sub-nets are further added to specifically handle each target. In sharp contrast, we attempt to learn a `universal' feature that preserves enough information for both motion and affinity models, which essentially simplifies the training and testing procedures.

\noindent\textbf{Object Affinity Models in Online MOT.}~In the data association phase, the object affinity model is usually used to link tracklets or detections cross frames in terms of the pairwise affinity, which is a crucial way to handle occlusions in online MOT. To produce reliable affinity estimations, object appearance {{cues are}} indispensable, and Siamese or triplet networks~\cite{chopra2005learning,lu2019see,wang2014learning} with metric learning provide powerful tools to acquire a discriminative and robust feature embedding. In particular, Leal-Taix{\'e} \etal~\cite{leal2016learning} apply a Siamese network to estimate the affinity of the provided detections by aggregating targets appearance and optical flow information. Son \etal~\cite{son2017multi} propose a quadruplet loss to stress targets appearance together with their temporal adjacencies. In~\cite{tang2017multiple}, a Siamese network is used to leverage human pose information for long-range target-relation modeling. Voigtlaender \etal~\cite{voigtlaender2019mots} extend the Mask R-CNN~\cite{he2017mask} with 3D convolutional layers and propose an association head to extract embedding vectors for each region proposals by using the batch hard triplet loss~\cite{hermans2017defense}. Bergmann \etal\cite{bergmann2019tracking} also present a short-term re-identification model based on the Siamese network. Xu \etal~\cite{xu2019spatial} jointly utilize the appearance, location and topology information to compute the affinity by employing the relation networks~\cite{wang2018non} in both the spatial and temporal domains. Notably, all these methods work on the tracking-by-detection mode. Differently, we in-depth inject metric learning into a SOT model, through a unified triplet network. It learns a discriminative feature for both the object motion prediction and affinity measure sub-tasks, bringing an effective yet efficient solution.


\section{Our Algorithm}

In this section, we first give a brief review of the Siamese SOT~\cite{bertinetto2016fully} (\S\ref{subsec:SiameseSOT}), as it is used as the backbone of our model. Then, the details of our \ourmodel~model are presented in \S\ref{subsec:ourmodel}. Finally, in \S\ref{subsec:pipeline}, we elaborate on our whole online MOT pipeline. As \ourmodel~utilizes a \textit{single} feature extraction network for both SOT based tracklet generation and object affinity measure, it presents a more efficient online solution with many non-trivial technical improvements.

\subsection{Preliminary of Siamese SOT}
\label{subsec:SiameseSOT}

Our backbone model is a recently proposed deep tracker: SiamFC~\cite{bertinetto2016fully}, which is based on a Siamese network and shows promising performance in the single object tracking field. It operates at around 120 fps on one GPU, built upon the lightweight AlexNet~\cite{krizhevsky2012imagenet}.

\begin{figure}[t]
  \centering
      \includegraphics[width=1 \linewidth]{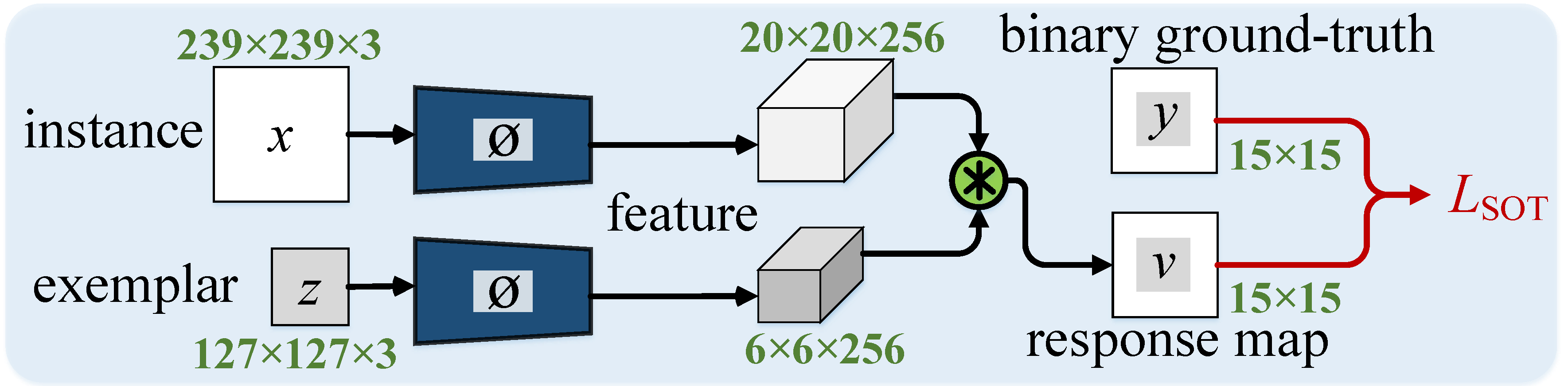}
\vspace{-15pt}
\captionsetup{font=small}
\caption{\small \textbf{Illustration of the network architecture of the Siamese SOT} during the training phase.}
\label{fig:SiamFC}
\vspace{-5mm}
\end{figure}

Basically, SiamFC transfers tracking task to patches matching in an embedding space. The Siamese network is learnt as the matching function, which is applied to find the most similar patch in the new frame compared with the initial target patch in the first frame. Specifically, as shown in Fig.~\ref{fig:SiamFC}, the Siamese tracker comprises two parameter-sharing branches, each of which is a 5-layer convolutional network $\phi$. One branch takes as input the target detection given at the first frame, called as exemplar. The other one takes as input the instance, \ie, the searching region in each subsequent frame including the candidate patches. Given the features embeddings: $\phi(z)$ and $\phi(x)$, of the exemplar $z$ and instance $x$, a cross correlation layer $\tau$ is applied to compare their similarity and obtain a response map $v$:
\begin{equation}\small
v = \tau(x, z) = \phi(x)*\phi(z)+b,
\label{eq:1}
\end{equation}
where `$*$' indicates the convolutional operator and $b$ is the biases term. Then, given a ground-truth map $y$, a logistic loss is applied on $v$ for training:
\begin{equation}\small
L_{\text{SOT}}=\sum\nolimits_{p\in \mathcal{P}}{\frac{1}{\vert \mathcal{P} \vert} \log{(1+e^{-v_p y_p})}},
\label{eq:sot}
\end{equation}
where $p$ indicates a candidate position in the lattice $\mathcal{P}$ of $x$.
For each candidate $x_p\!\in\!x$ from the instance input $x$, $v_p$ is the response value of an exemplar-candidate pair, \ie, $v_p\!=\!f(x_p, z)$ , and $y_p\!\in\!\{+1, -1\}$ is the corresponding ground-truth label for $v_p$.

\begin{figure*}[t]
  \centering
      \includegraphics[width=0.99 \linewidth]{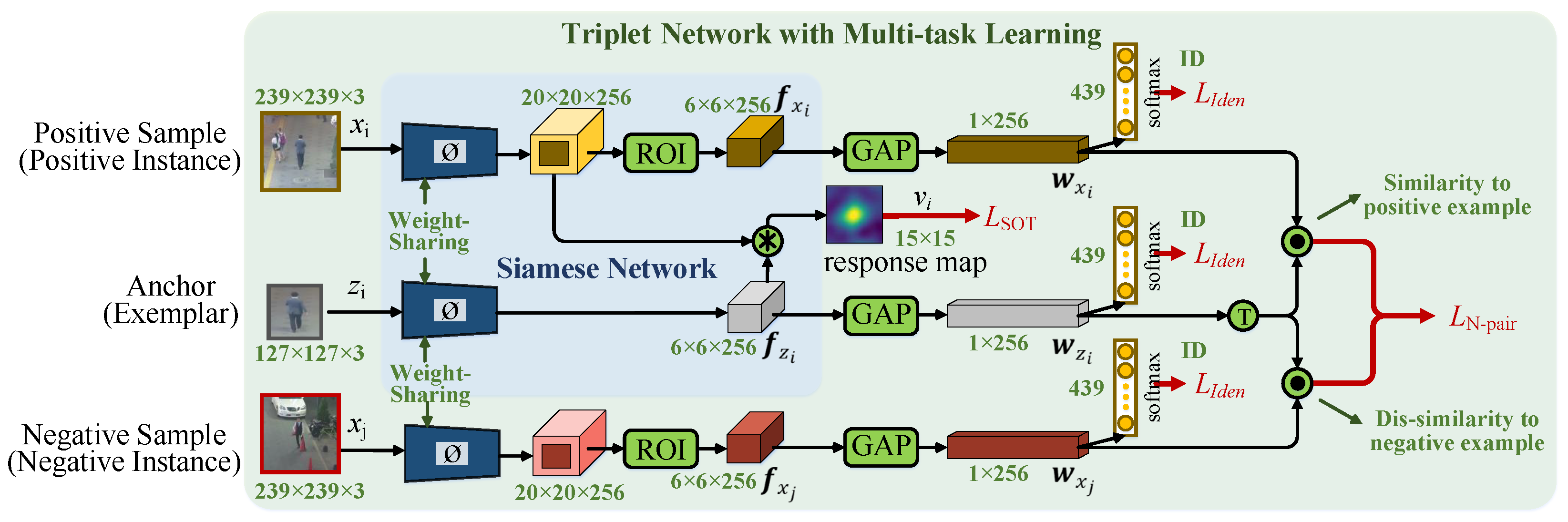}
\vspace{-8pt}
\captionsetup{font=small}
\caption{\small \textbf{Illustration of our proposed \ourmodel~model}, which is built upon a triplet architecture with multi-task learning. \ourmodel~simultaneously learns two tasks: SOT based object motion prediction and affinity-dependent ranking, producing a strong feature that is applicable to both the tracklet generation as well as the affinity measure phases.}
\label{fig:dualtask}
\vspace{-3mm}
\end{figure*}

\subsection{Our \ourmodel~Model for Online MOT}
\label{subsec:ourmodel}


\noindent\textbf{Main Idea.}
Previous SOT based online MOT methods typically design an extra network for affinity measure, in addition to the SOT network.
In contrast, we try to integrate the object motion and affinity networks into a unified model. This brings several advantages, as mentioned in \S\ref{sec:intro}.
The core idea is to enforce the network to simultaneously learn the two tasks: single object tracking and affinity prediction, forming a unified multi-task learning framework. Some ones may concern the features obtained from top-performing SOTs are already good enough for affinity measure. Actually, though SOT features are powerful, they are not discriminative enough to estimate a reliable affinity. This is because SOTs rarely access the identity information during training, thus their features typically distinguish targets from the substantial background well, while capture
relatively less identity information. From the perspective of data association, SOT features have already encoded some useful information, thus it is more desirable and efficient to make use of these features instead of learning extra `affinity features' from scratch. These considerations motivate us to learn a unified yet powerful feature that is applicable to both tasks, yielding an elegant online MOT framework.

\noindent\textbf{Triplet-based MOT Framework.}~To achieve our goal, our \ourmodel~model is designed as a triplet network architecture, as shown in Fig.~\ref{fig:dualtask}, where the triplet network comprises three weight-sharing branches, \ie, an exemplar branch, a positive-instance branch and a negative-instance branch. 
We adopt the exemplar as the \textit{anchor}. The instances from the same targets are used as \textit{positive samples}, while the ones from different targets as \textit{negative}. The integration of the exemplar branch and positive-instance branch can be viewed as a Siamese tracking network, while the whole triplet network yields a unified metric learning framework.

Specifically, for the $i^{th}$ target, given an exemplar $z_i$, a positive-instance $x_i$, and a negative-instance $x_j$ sampled from a different target $j$, we extract their features from the backbone AlexNet: $\bm{f}_{z_i}\!=\!\phi(z_i)\!\in\!\mathbb{R}^{6\times6\times256}$, $\bm{f}_{x_i}\!=\!\phi(x_i)\!\in\!\mathbb{R}^{20\times20\times256}$, and $\bm{f}_{x_j}\!=\!\phi(x_j)\!\in\!\mathbb{R}^{20\times20\times256}$. Then, for the single object tracking task, it can be trained over $(\bm{f}_{z_i}, \bm{f}_{x_i})$ using Eq.~\ref{eq:sot}.

For the whole triplet-based model, it is designed to learn the ranking task for affinity estimation. This is achieved by a metric learning paradigm, \ie, enforcing the features of positive examples closer to the anchor than other negative examples. Specifically, we first apply the ROI-Align~\cite{he2017mask} layer on $\bm{f}_{x_i}$ and $\bm{f}_{x_j}$ respectively, to extract two $6\!\times\!6\!\times\!256$ target features from the centers of $x_i$ and $x_j$ (as the targets are centred at the instance examples during training~\cite{bertinetto2016fully}).
Such operation allows the model to specifically focus on learning more identity-discriminative features for affinity measure, suppress the information from cluster background, and produce feature maps with the same resolution to the anchor feature $\bm{f}_{z_i}$. Then a global
average pooling (GAP) is applied to the anchor feature $\bm{f}_{z_i}$, {{as well as}} the aligned features of $\bm{f}_{x_i}$ and $\bm{f}_{x_j}$, producing three 256-\textit{d} features, denoted as $\bm{w}_{z_i}$, $\bm{w}_{x_i}$ and $\bm{w}_{x_j}$, respectively. This enforces the regularization of the network and reduces model size.

Given a mini-batch of $N$ training sample pairs, \eg, $\mathcal{B}\!=\!\{(x_i,z_i)\}_{i=1}^N$, a standard triplet loss~\cite{weinberger2006distance} works in the following format:
\begin{equation}\small
\begin{aligned}
\!\!L_{\text{Tri}}\!=\!\frac{1}{N}\!\sum^N\nolimits_{i,j}\max(0, ||\bm{w}_{z_i}\!\!-\!\bm{w}_{x_i}\!||_2^2\!-\!||\bm{w}_{z_i}\!\!-\!\bm{w}_{x_j}\!||_2^2\!+\!m),\!\!
\end{aligned}
\label{eq:tri_loss}
\end{equation}
where $m$ is a margin that is enforced between positive and negative pairs. The objective of this loss is to keep distance between the anchor and positive smaller than the distance between the anchor and negative. However, in our batch construction, the number of the positive samples is significantly smaller than the negative ones, which will restrict the performance of the triplet loss $L_{\text{Tri}}$ in hard data mining~\cite{hermans2017defense}. To overcome this hurdle, we replace Eq.~\ref{eq:tri_loss} with a $N$-pair loss~\cite{sohn2016improved}:
\begin{equation}\small
\label{eq:npair_loss}
\begin{aligned}
\!\!L_{\text{N-pair}}\!=\!\!\frac{1}{N}\!\!\sum\nolimits_{i=1}^{N}{\!\log\!{\Big(\!1\!+\!\!\sum^N\nolimits_{i\neq j}{\!\exp{\!\big(\bm{w}_{z_i}^\top\bm{w}_{x_j}
\!-\!\bm{w}_{z_i}^\top\bm{w}_{x_i}\big)\!\Big)}}}}.\!\!
\end{aligned}
\end{equation}
The rationale is that, after looping over all the triplets in $\mathcal{B}$, the final distance metric can be balanced correctly.

Additionally, with the target identity at hand, we can further minimize a cross-entropy based identification loss~\cite{chu2019online}:
\vspace{-6pt}
\begin{equation}\small
\begin{aligned}
\label{eq:iden_loss}
L_{\text{Iden}}=-(\frac{1}{N}\sum\nolimits_{i=1}^{N}{\log{\hat{p}_{z_i}}}+\frac{1}{N}\sum\nolimits_{i=1}^{N}{\log{\hat{p}_{x_i}}}),
\vspace{-2pt}
\end{aligned}
\end{equation}
where $\hat{p}_{z_i}\!\in\![0,1]$ is the predicted probability of $z_i$ for the i$^{th}$ identity class. The identity prediction score is obtained by applying two fully connected layers (with dimension 512 and 439) and a \textit{softmax} layer over $\bm{w}_{z_i}$ or $\bm{w}_{x_i}$.
 Please note that there are a total of 439 identities in our training set.

Hence, the final loss is computed as a combination of the SOT loss $L_{\text{SOT}}$, defined over the Siamese network, and the affinity-related losses $L_{\text{N-pair}}$ and $L_{\text{Iden}}$, defined over the whole triplet network:\!\!
\vspace{-2pt}
\begin{equation}\small
L=L_{\text{SOT}}+(\lambda_1L_{\text{N-pair}}+\lambda_2L_{\text{Iden}}),
\label{eq:overall}
\vspace{-2pt}
\end{equation}
where $\lambda$s are the coefficients to balance different losses. In this way, both the SOT based motion model and the ranking based affinity model can be trained end-to-end in a unified triplet network, which facilitates the training process.

\begin{figure}[t]
  \centering
      \includegraphics[width=1 \linewidth]{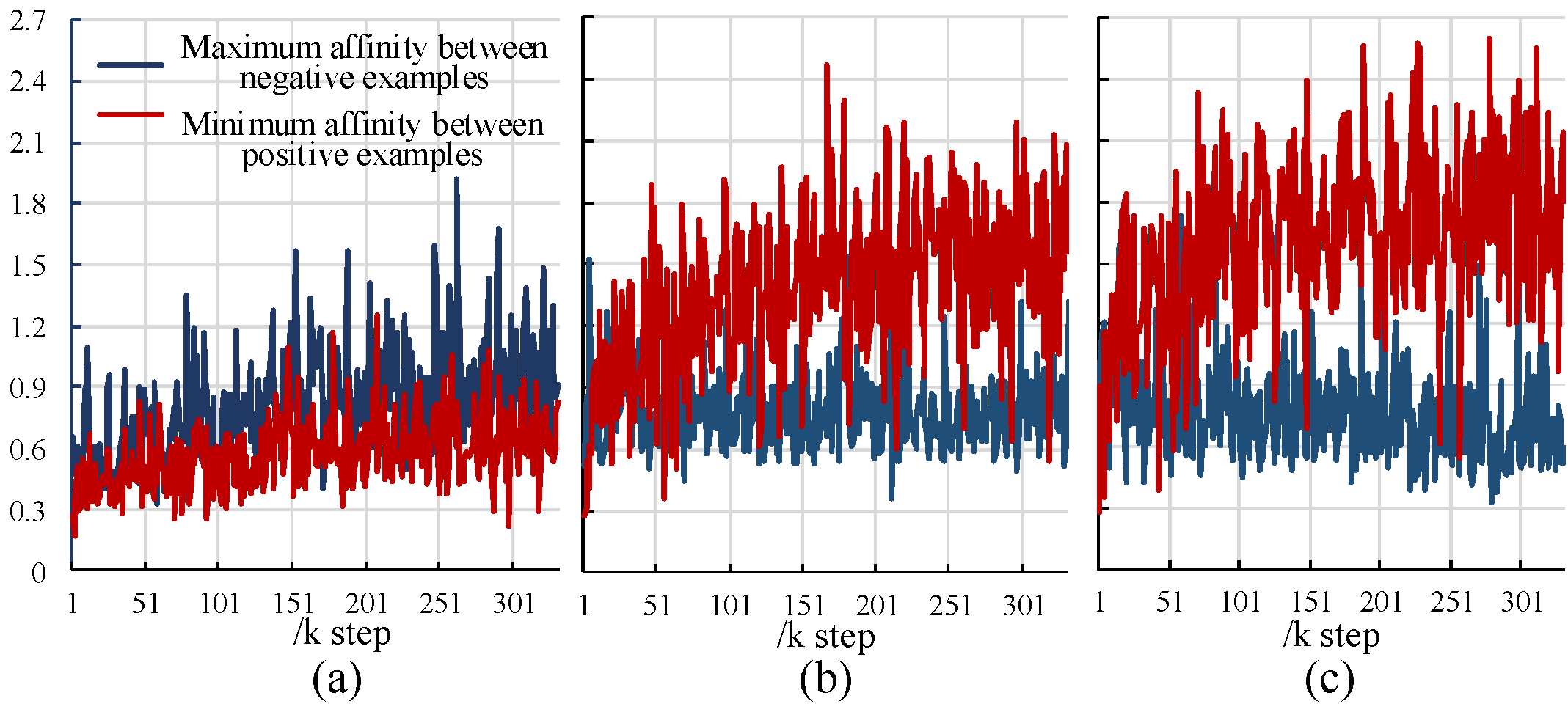}
\vspace{-18pt}
\captionsetup{font=small}
\caption{\small \textbf{Affinities measured using the features} (a) $\bm{f}$ from the Siamese SOT with $L_{\text{SOT}}$ loss, (b) $\bm{w}$ from the triplet network with multi-task learning, and (c) $\bm{w}^{\text{AFF}}$ from our whole \ourmodel~with multi-task learning and TSA module, respectively.}
\vspace{-12pt}
\label{fig:distance}
\end{figure}

Furthermore, with our multi-task design, we can derive a reliable affinity from the features extracted from our model:\!\!\!
\vspace{-2pt}
\begin{equation}\small
\begin{aligned}
c = \bm{w}_{I}^\top\bm{w}_{I'},
\label{eq:affinity}
\end{aligned}
\vspace{-2pt}
\end{equation}
where $I$ and $I'$ are two image patch inputs, \eg, an exemplar with an instance region or a detection patch, and $c$ is the affinity. To in-depth analyze the advantage of the features learnt from our model in affinity measure, we use the features extracted from the Siamese SOT model to compute the affinity, where the SOT model does not use either the additional branch or the extra losses (\ie, $L_{\text{N-pair}}$ and $L_{\text{Iden}}$). Fig.~\ref{fig:distance} gives the performance comparison of the two models with hard cases, \eg, the affinity between negative sample pairs with similar appearance or that {{between}} positive sample pairs with changeable appearance. From Fig.~\ref{fig:distance}~(a) we can find that, when only using $L_{\text{SOT}}$, the affinity between negative sample pairs is even larger than the one between positive sample pairs. This clearly demonstrates the weak discriminability of the SOT features. In Fig.~\ref{fig:distance}~(b) {{and (c)}}, the affinity between positive sample pairs is substantially larger than that between negative ones even in hard cases, which proves our multi-task features $\bm{w}$ are highly applicable for affinity measure. More detailed quantitative experiments can be found at \S\ref{sec:experiments}.

\noindent\textbf{Task-Specific Attention Module.} For our triplet-based model described above, an identical feature produced by the backbone AlexNet $\phi(\cdot)$ is used for both the SOT based motion prediction and affinity measure tasks. Potential problems of such design lie in the loss of sensibility to
subtle distinctions between the two tasks and the ignorance of their task-specific factors. A meaningful feature for SOT may not best fit affinity measure, vice verse. For example, context information is often stressed in SOT, \eg, auxiliary objects approaching the target may afford correlation information for tracking~\cite{yang2009context,sohn2016improved}. However, for affinity measure, local semantic features around the key points are more informative to identify the query target~\cite{cheng2016person, Su2017pose}, while the auxiliary objects may interfere with the determination. To address this issue, we further equip our model with a task-specific attention (TSA) module for emphasizing task-aware feature learning with very low computation cost.

Our TSA module is designed based on the famous Squeeze-and-Excitation Network (SENet)~\cite{hu2017squeeze}, as it does not rely on extra input and is trivial in runtime, which are essential for online MOT. It re-weights the feature response across channels using the \textit{squeeze} and \textit{excitation} operations. More specifically, the \textit{squeeze} operator acquires global information via aggregating the features across all the spatial locations through channel-wise global average pooling:
\vspace{-2pt}
\begin{equation}\small
s_l = \text{GAP}_l(\bm{f}) = \text{GAP}_l(\phi(\cdot))\in \mathbb{R},
\vspace{-2pt}
\end{equation}
where $\text{GAP}_l$ indicates global average pooling over the feature $\bm{f}$ in $l^{th}$ channel.
In the \textit{excite} step, a gating
mechanism is employed on the channel-wise descriptor $\bm{s} = [s_1,s_2,\dots,s_{256}]\!\in\!\mathbb{R}^{256}$:
\vspace{-2pt}
\begin{equation}\small
\bm{a} = \sigma(\bm{W}_2\delta(\bm{W}_1\bm{s})) = [a_1,a_2,\dots,a_{256}]\in [0,1]^{256}.
\vspace{-2pt}
\end{equation}
$\sigma$ and $\delta$ are \textit{sigmoid} and \textit{ReLU} functions, respectively. Through the dimensionality reduction and
increasing operations (parameterized by two fully connected layers $\bm{W}_1\!\in\!\mathbb{R}^{64\times256}$ and $\bm{W}_2\!\in\!\mathbb{R}^{256\times64}$), the attention vector $\bm{a}$ encodes non-mutually-exclusive
relations among the 256 channels.

\begin{figure}[t]
  \centering
      \includegraphics[width=1 \linewidth]{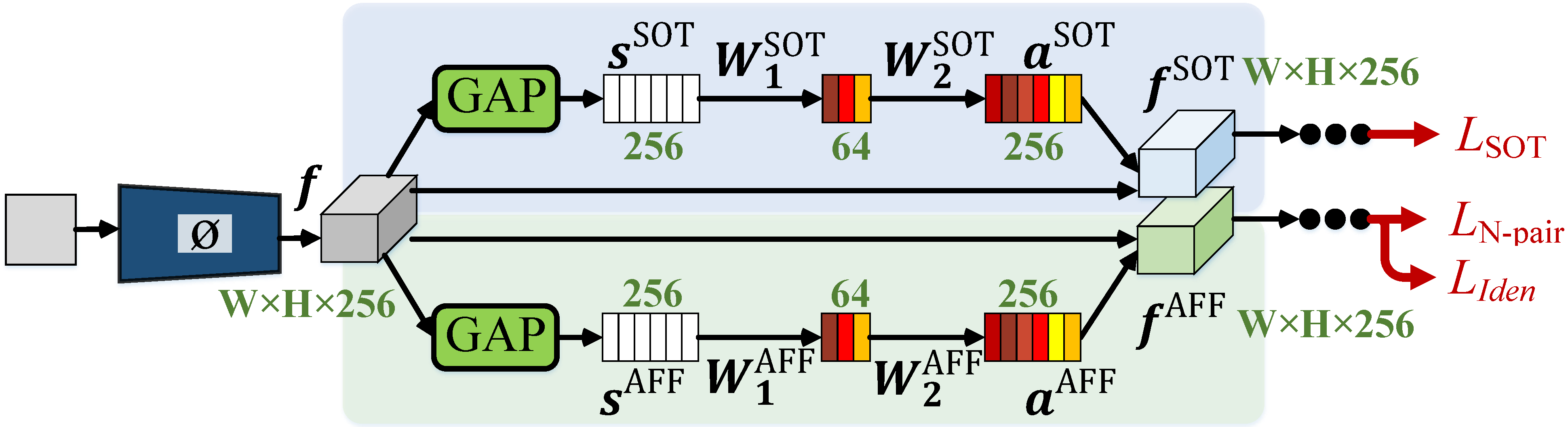}
\vspace{-16pt}
\captionsetup{font=small}
\caption{\small \textbf{Illustration of the TSA module}, which enables our model to stress task-specific features.}
\label{fig:TSA}
\vspace{-8pt}
\end{figure}
\begin{figure}[t]
  \centering
      \includegraphics[width=0.99 \linewidth]{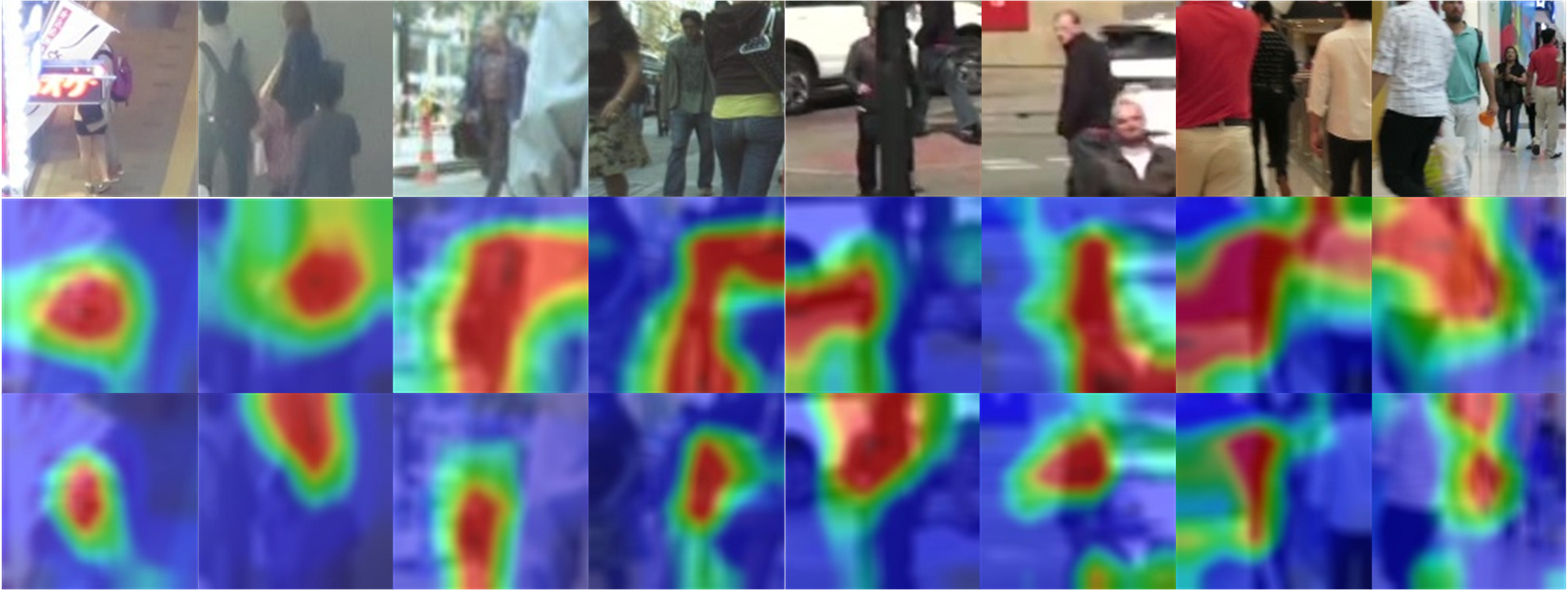}
\vspace{-6pt}
\captionsetup{font=small}
\caption{\small \textbf{Visualization of task-specific features} stressed by TSA module. Features extracted for {tracking part} are shown in the 2$^{nd}$ row, while those for affinity measure part are in the 3$^{rd}$ row.}
\label{fig:attention}
\vspace{-12pt}
\end{figure}

Using the SENet framework, our TSA module learns two kinds of attentions: $\bm{a}^{\text{SOT}}$ and $\bm{a}^{\text{AFF}}$ for addressing different tasks (see Fig.~\ref{fig:TSA}). $\bm{a}^{\text{SOT}}$ and $\bm{a}^{\text{AFF}}$ are first applied to re-weight the channels of
the `universal' feature $\bm{f}\!=\![\bm{f}_{1},\dots, \bm{f}_{256}]$ extracted from the backbone AlexNet:
\vspace{-4pt}
\begin{equation}\small
\begin{aligned}
\bm{f}^{\text{SOT}}& = [a^{\text{SOT}}_1\cdot\bm{f}_{1},\dots, a^{\text{SOT}}_{256}\cdot\bm{f}_{256}],\\
\bm{f}^{\text{AFF}}& = [a^{\text{AFF}}_1\cdot\bm{f}_{1},\dots, a^{\text{AFF}}_{256}\cdot\bm{f}_{256}].\\
\end{aligned}
\end{equation}

Then we feed the supervision of $L_{\text{SOT}}$ to the SOT-aware feature $\bm{f}^{\text{SOT}}$, while add $L_{\text{N-pair}}$ and $L_{\text{Iden}}$ losses to the affinity-related feature $\bm{w}^{\text{AFF}}$ (derived from $\bm{f}^{\text{AFF}}$, as described before). In this way, the TSA module will learn to generate task-specific attentions. Through our lightweight TSA mechanism, our model is able to produce task-specific features while using a same backbone network $\phi(\cdot)$. For the single object tracking task, the SOT-aware attention $\bm{a}^{\text{SOT}}$ can stress useful context for boosting tracking accuracy. For the affinity measure, the affinity-aware attention $\bm{a}^{\text{AFF}}$ is employed to capture fine-grained local semantic features. Thus the targets with changeable appearance can be better aligned. From Fig.~\ref{fig:distance} (c), we can observe further improved affinity estimations using the affinity-specific attention enhanced feature $\bm{w}^{\text{AFF}}$. Visualization of the attention-enhanced features for each task is presented in Fig.~\ref{fig:attention}. More detailed quantitative analyses can be found in \S\ref{sec:ablation}.

\subsection{Our Online MOT Pipeline}
\label{subsec:pipeline}

We have elaborated on our network architecture. Next we will detail our whole pipeline for online MOT. Basically, each target is associated with two states, \ie, tracked or occluded, decided by the occlusion detection. We first apply our \ourmodel~ (working on the SOT mode) to generate tracklets for the tracked targets. Then we perform data association based on the affinity produced by \ourmodel~(working on the ranking mode), to recover the occluded targets.

\noindent\textbf{Tracklet Generation and Occlusion Detection:} During tracking, our \ourmodel~is applied to each target (exemplar $z$), which is initialized by the provided detections. \ourmodel~is able to update the position of each target used as a SOT (relying on the SOT-specific feature $\bm{f}^{\text{SOT}}$). Simultaneously, it measures the affinity between the exemplar and instances in subsequent frames to detect occlusions (using the affinity estimation-related feature $\bm{f}^{\text{AFF}}$).

Concretely, we use a search region centred at the previous position of the target as the instance $x$. Given $z$ and $x$, we get a response map $v$ via Eq.~\ref{eq:1} with SOT-specific feature $\bm{f}^{\text{SOT}}$. Then the target bounding box (bbox) is obtained according to the position with the maximum score in $v$~\cite{bertinetto2016fully}.  

\begin{figure}
\begin{center}
\includegraphics[width=0.49\textwidth]{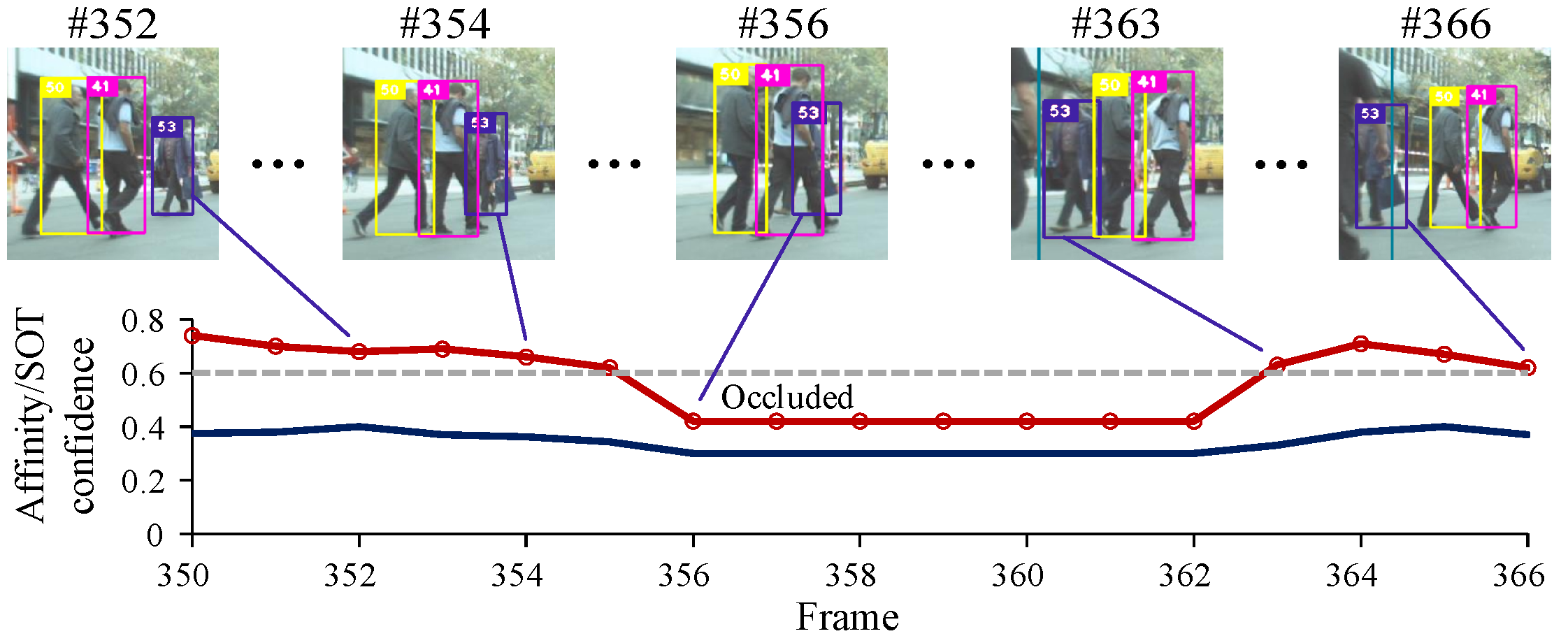}
\end{center}
\vspace{-18pt}
\captionsetup{font=small}
\caption{\small {\textbf{Illustration of the occlusion handling.} The red line denotes the affinity produced by our \ourmodel~model, while the blue line signifies the confidence of the Siamese SOT. Our proposed model is more robust in detecting and addressing the occlusions.}}\label{fig:occlusion}
\vspace{-9pt}
\end{figure}

Meanwhile, with the exemplar $z$ and instance $x$, UMA computes the affinity for detecting occlusions. It works in the ranking mode and uses the affinity-specific features $\bm{w}_z^{\text{AFF}\!\!}$ and $\bm{w}_x^{\text{AFF}\!\!}$ to get the affinity $c$ (Eq.~\ref{eq:affinity}).
Note that the target may appear in any part of $x$ during the tracking stage, thus we apply ROI-Align~\cite{he2017mask} on the instance feature ($\bm{f}_{x}^{\text{AFF}\!}\!\in\!\mathbb{R}^{22\times22\times256}$ during testing) to obtain an aligned target feature, with the bbox provided by the SOT.
Then we get $\bm{w}_x^{\text{AFF}}$ through GAP and further compute the affinity $c$. Compared with previous works~\cite{xiang2015learning,sadeghian2017tracking,zhu2018online} that use the confidence produced by SOTs to detect the occlusions, our method gives a more robust result, which is illustrated in Fig.~\ref{fig:occlusion}.
Additionally, following~\cite{zhu2018online}, we integrate the affinity with the historic average intersection-over-union (IOU) between the tracklet and the nearest detection, for filtering out the FP tracking results and detecting occlusions more reliably.
Once the affinity $c$ is below a threshold $\alpha$ or the average IOU is below $\beta$, the target is recognized to be occluded; tracked otherwise. We further refine the tracked bboxes by averaging the nearest detection with greedy algorithm. Then the refined bboxes are gathered as the tracklet of the target $z$. Detections that have IOU below a certain threshold $\gamma$ with any tracking bboxes will be regarded as candidate detections, \eg, a reappearing occluded target or an entirely new target.

\begin{table*}
\centering\small
\setlength\tabcolsep{4pt}
\renewcommand\arraystretch{1.05}
\resizebox{0.99\textwidth}{!}{
\begin{tabular}{c|l||cc|ccccccccc}
\hline
Mode    &~~~~~~~~~Method     &Publication &Year   & \multicolumn{1}{c}{MOTA$\uparrow$}   & \multicolumn{1}{c}{IDF1$\uparrow$}
& \multicolumn{1}{c}{MOTP $\uparrow$}   & \multicolumn{1}{c}{~~~MT $\uparrow$~~~}        & \multicolumn{1}{c}{~~~ML $\downarrow$~~~}        & \multicolumn{1}{c}{~~~FP $\downarrow$~~~}      & \multicolumn{1}{c}{~~~FN $\downarrow$~~~}       & \multicolumn{1}{c}{~~IDS $\downarrow$~~}     & \multicolumn{1}{c}{~~Hz $\uparrow$} \\\hline
        & STAM~\cite{chu2017online}     &ICCV&2017     & 46.0         &50.0            & 74.9                        & 14.60\%                        & 43.60\%                        & 6,895                        & 91,117                        & {\color{red} \textbf{473}}
        & 0.2 \\
        & AMIR~\cite{sadeghian2017tracking}    &ICCV&2017      & 47.2       &46.3  & 75.8 & 14.00\%                        & 41.60\%                        & {\color{red} \textbf{2,681}} & 92,856                        & 774                             & 1.0              \\
 & DMAN~\cite{zhu2018online}     &ECCV&2018   & 46.1      &{\color{red} \textbf{54.8}}        & 73.8                        & 17.40\%                        & 42.70\%                        & 7,909                        & 89,874                        & 532                                & 0.3           \\

     Online  & C-DRL~\cite{ren2018collaborative}   &ECCV&2018      & 47.3       & -         & 74.6                        & 17.40\%                      & 39.90\%                        & 6,375                         & 88,543                                                  & -        & 1.0                    \\
        & KCF16~\cite{chu2019online}   &WACV&2019      & 48.8       &47.2         & 75.7                        & 15.80\%                        & 38.10\%                        & 5,875                         & 86,567                         & 906                               & 0.1                    \\
        & Tracktor++~\cite{bergmann2019tracking}   &ICCV&2019      & {\color{red} \textbf{54.4}}       &52.5         & {\color{red} \textbf{78.2}}                        & {\color{red} \textbf{19.00\%}}                        & 36.90\%                        & 3,280                         & {\color{red} \textbf{79,149}}                         & 682                               & 2.0                    \\
        & \textbf{\ourmodel~(ours)} &CVPR&2020  & 50.5 & 52.8 & 74.1                        & 17.80\% & {\color{red} \textbf{33.70\%}} & 7,587                        & 81,924  & 685                              & {\color{red} \textbf{5.0}}        \\ \hline
         & QuadMOT~\cite{son2017multi}  &CVPR&2017     & 44.1       &38.3           & 76.4                        & 14.60\%                        & 44.90\%                        & {\color{blue} \textbf{6,388}}                        & 94,775                        & 745                              & 1.8           \\
         & FWT~\cite{Henschel2018Fusion}    &CVPRW &2018                   & 47.8 &47.8                      & {\color{blue} \textbf{77.0}}                        & 19.10\%                        & 38.20\%                       & 8,886                        & 85,487                        & 852  & 0.2                    \\
    Offline   & MHTBLSTM~\cite{kim2018multi}  &ECCV&2018     & 42.1      &47.8        & 75.9                        & 14.90\%                        & 44.40\%                        & 11,637                        & 93,172                        & 753                              & 1.8           \\

& JCC~\cite{keuper2018motion}    &TPAMI &2018                   & 47.1 &52.3                      & -                        & {\color{blue} \textbf{20.40\%}}                       & {46.90\%}                        & 6,703                        & 89,368                        & {\color{blue} \textbf{370}}  & 1.8                    \\
&TLMHT~\cite{sheng2018iterative}   &TCSVT&2018     & {\color{blue} \textbf{48.7}} & {\color{blue} \textbf{55.3}}                        &   76.4                      & 15.70\%                        & 44.50\%                        & 6,632                        & 86,504                        & 413                             & {\color{blue} \textbf{4.8}}
			 \\
& LNUH~\cite{wen2018learning}
            &AAAI&2019     & 47.5 & 43.6                        & -                        & 19.40\%                         & {\color{blue} \textbf{36.90\%}}                        & 13,002                        & {\color{blue} \textbf{81,762}}                        & 1,035                         & 0.8
          	  \\
          	
 \hline
\end{tabular}%
}
\vspace{-6pt}
\captionsetup{font=small}
\caption{\small \label{tb:mot16}\textbf{Quantitative results on MOT16.} The best scores of online and offline MOT methods are marked in {\color{red} \textbf{red}} and {\color{blue} \textbf{blue}}, respectively. }
\vspace{-8pt}
\label{tb:mot16}
\end{table*}

\noindent\textbf{Data Association:} During data association, we deal with those candidate detections and address the occluded targets, \ie, recognizing a candidate detection as a reappearing occluded or an entirely new target, and then recovering its identity (if the first case) or assigning a new identity (if the second).
Different from prior work designing complicated strategies~\cite{tang2017multiple,dehghan2015gmmcp,chari2015pairwise}, we use a relatively simple data association method, due to the reliable affinity measured from our \ourmodel. 
Given the candidate detection set $\mathcal{D}$ and tracklet set $\mathcal{T}$ of occluded targets, produced from the previous stage, we build an affinity matrix $C\!\in\!\mathbb{R}^{|\mathcal{D}|\times|\mathcal{T}|}$ to obtain the optimal assignment. More specifically, for a tracklet $T\!\in\!\mathcal{T}$, we uniformly sample $K$ samples from $T$, \ie, $\{t_1,t_2,...,t_K\}$. Then the affinity between $T$ and a candidate detection $d\!\in\!\mathcal{D}$ is calculated by:
\vspace{-4pt}
\begin{equation}\small
 c'=\frac{1}{K}\sum\nolimits_{k=1}^{K}\bm{w}_{d}^\top\bm{w}_{t_k}.
\end{equation}
After computing all the affinity, we construct the cost matrix $C$ (affinity matrix) and obtain the optimal assignment by applying the Hungarian algorithm~\cite{kuhn1955hungarian} over $C$.
According to the assignment result, a candidate detection is assigned the identity of an occluded target that links to it. If a candidate detection does not link to any occluded targets, we view it as an entirely new target and assign it a new identity.

\noindent\textbf{Trajectory Management:} {For trajectory initialization, we adopt method in~\cite{liu2019multiple} to alleviate the influence caused by FP detections. Besides, a target will be terminated if it moves out of the view or keeps occluded for over certain frames.}

\vspace{-1pt}
\section{Experiments}
\label{sec:experiments}
\vspace{-1pt}
\label{sec:datasets}
\vspace{-2pt}
\noindent\textbf{Datasets:} We evaluate our approach on the MOT16 and MOT17 datasets from MOT Challenge~\cite{milan2016mot16}, which is a standardized benchmark focusing on multiple people tracking. MOT16 dataset contains 14 video sequences (7 for training and 7 for testing) from unconstrained environments filmed with both static and moving cameras. It provides ground-truth annotations for the training set and detections~\cite{felzenszwalb2010object} for both sets.
MOT17 contains more video sequences than MOT16, and provides accurate annotations and richer detections from different detectors, \ie, DPM~\cite{felzenszwalb2010object}, SDP~\cite{yang2016exploit} and FRCNN~\cite{ren2015faster}. For the evaluation of the two test sets, the results are submitted to the server of the benchmarks.


\noindent\textbf{Evaluation Metrics:} For quantitative performance evaluation, we adopt the widely used CLEAR MOT metrics~\textit{et al.}~\cite{bernardin2008evaluating}, \ie, the multiple object tracking accuracy (MOTA), multiple object tracking precision (MOTP), false positives (FP), false negatives (FN), identity switches (IDS) and IDF1 score. In addition, the metrics defined in~\cite{li2009learning} are also used, including the percentage of mostly
tracked targets (MT) and the percentage of mostly lost targets (ML). MT refers to the ratio of ground-truth trajectories that are covered by any track hypothesis for at least 80\% of their respective life span. ML is computed as the ratio of ground-truth trajectories that are covered by any track hypothesis for at most 20\% of their respective life span.

\noindent\textbf{Implementation Details:}
We adopt the sequences in MOT17 for training. The exemplar-positive instance pair $<$$x_i, z_i$$>$ is composed of image patches from the same targets in various frames. Patches from different targets are then chosen as the negative instances. During training, the sizes of the exemplar and instance are set as $127\!\times\!127$ and $239\!\times\!239$, respectively. The AlexNet pre-trained model on ImageNet dataset~\cite{deng2009imagenet} is used to initiate the shared part of our \ourmodel~model, while other layers are initialized through He initialization~\cite{he2015delving}. We use the learning rate configuration in~\cite{bertinetto2016fully}. The coefficient parameters in Eq.~\ref{eq:overall} are set as $\lambda_1\!=\!\lambda_2\!=\!0.1$. The total loss is minimized through momentum optimization~\cite{sutskever2013importance} with a mini-batch of size 8. The thresholds $\alpha$ and $\beta$ used for detecting occlusions are set to 0.6 and 0.5, respectively. The threshold $\gamma$ is set to 0.5, which decides whether a detection is selected as candidates for data association. {We empirically set the threshold for terminating an occluded target as 30 frames.}


\begin{figure}
\begin{center}
\includegraphics[width=0.96\linewidth]{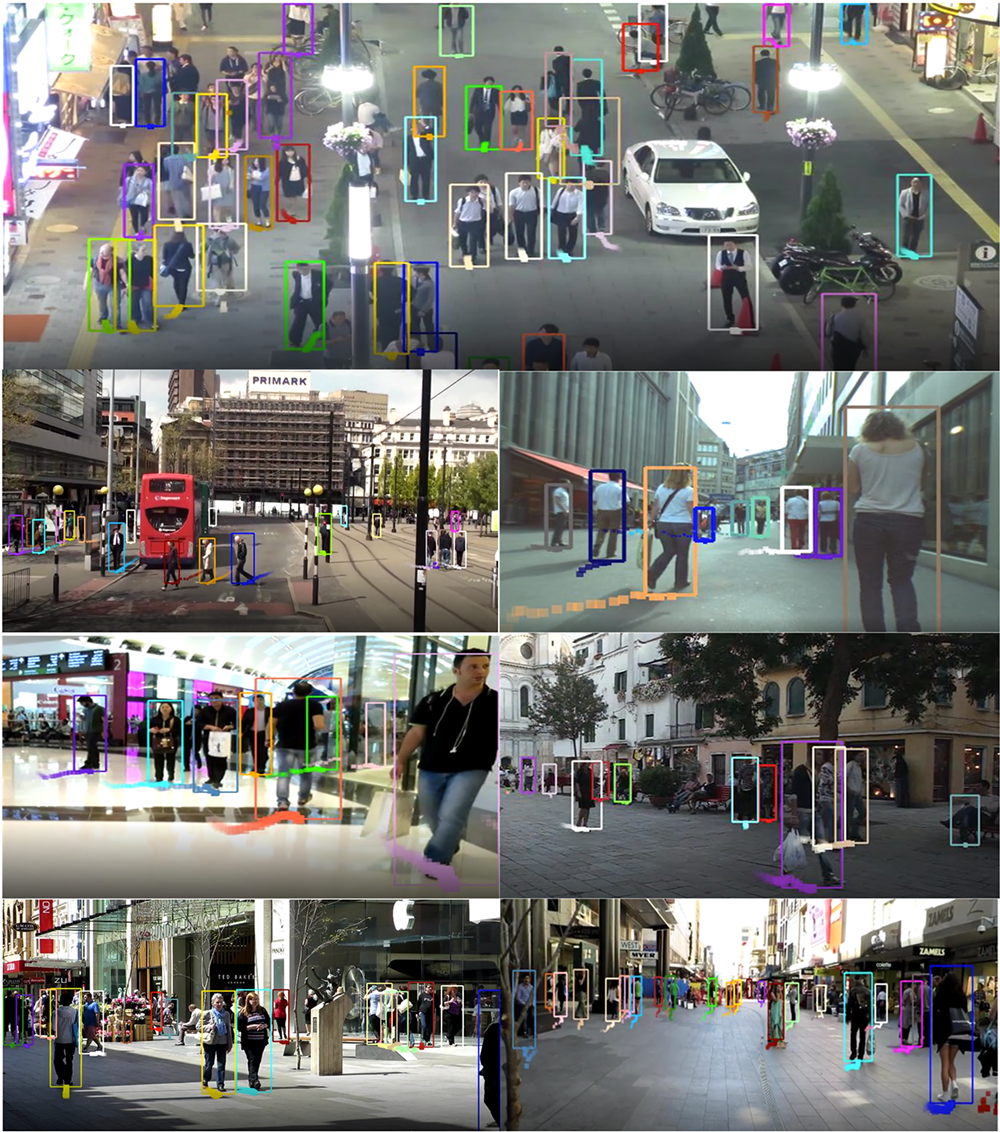}
\end{center}
\vspace{-15pt}
\captionsetup{font=small}
\caption{\small \textbf{Qualitative tracking results on the test sequences of MOT17 benchmark}. The color of each bounding box indicates the target identity. The dotted line under each bounding box denotes the recent tracklet of each target.}\label{fig:results}
\vspace{-12pt}
\end{figure}


\begin{table*}
\centering\small
\setlength\tabcolsep{4pt}
\renewcommand\arraystretch{1.05}
\resizebox{0.99\textwidth}{!}{
\begin{tabular}{c|l||cc|cccccccccc}
\hline
Mode    &~~~~~~~~~Method     &Publication &Year    & \multicolumn{1}{c}{MOTA$\uparrow$}
& \multicolumn{1}{c}{IDF1$\uparrow$}  & \multicolumn{1}{c}{MOTP $\uparrow$}   & \multicolumn{1}{c}{~~~MT $\uparrow$~~~}        & \multicolumn{1}{c}{~~~ML $\downarrow$~~~}        & \multicolumn{1}{c}{~~~FP $\downarrow$~~~}      & \multicolumn{1}{c}{~~~FN $\downarrow$~~~}       & \multicolumn{1}{c}{~~IDS $\downarrow$~~}    &  \multicolumn{1}{c}{~~Hz $\uparrow$}
    \\\hline
       & DMAN~\cite{zhu2018online}    &ECCV &2018                   & 48.2           & \color{red} \textbf{55.7}            & 75.5                        & 19.30\%                        & 38.30\%                        & 26,218                        & 263,608                        & 2,194  & 0.3 \\
       & MTDF~\cite{fu2019multi}   &TMM &2019              & 49.6   &45.2                     & 74.5                        & 18.90\%                         & 33.10\%                        & 37,124                        & 241,768                        & 5,567                                  & 1.2            \\
     Online   & FAMNet~\cite{chu2019famnet}   &ICCV&2019      & 52.0       & 48.7         & 76.5                        & 19.10\%                      & 33.40\%                        & 14,138                         & 253,616                                                  & 3,072        & 0.6                    \\
        & Tracktor++~\cite{bergmann2019tracking}    &ICCV &2019    & {\color{red} \textbf{53.5}} &52.3 & {\color{red} \textbf{78.0}}                        & 19.50\%   & 36.60\%   & {\color{red} \textbf{12,201}}                        & 248,047  & {\color{red} \textbf{2,072}}                                & 2.0             \\
        & \textbf{\ourmodel~(ours)}    &CVPR &2020    & {53.1} &54.4 & 75.5                        & {\color{red} \textbf{21.50\%}}   & {\color{red} \textbf{31.80\%}}   & 22,893                        & {\color{red} \textbf{239,534}}  & 2,251                                & {\color{red} \textbf{5.0}}             \\ \hline
        & EDMT~\cite{chen2017enhancing}   &CVPRW &2017                      & 50.0 &51.3 & 77.3                        & {\color{blue} \textbf{21.60\%}} & 36.30\% & 32,279                        & {\color{blue} \textbf{247,297}} & 2,264     &0.6                \\
       & FWT~\cite{Henschel2018Fusion}    &CVPRW &2018                   & {\color{blue} \textbf{51.3}} &47.6                      & 77.0                        & 21.40\%                        & {\color{blue} \textbf{35.20\%}}                        & 24,101                        & 247,921                        & 2,648  & 0.2                    \\
  Offline  & MOTBLSTM~\cite{kim2018multi}    &ECCV &2018                   & 47.5         &51.9              & 77.5                      & 18.20\%                        & 41.70\%                        & 25,981                        & 268,042                        & 2,069  & 1.9                        \\
    & TLMHT~\cite{sheng2018iterative}    &TCSVT &2018                   & 50.6              & 56.5         & {\color{blue} \textbf{77.6}}                        & 17.60\%                        & 43.40\%                        & {\color{blue} \textbf{22,213}}                        & 255,030                        & {\color{blue} \textbf{1,407}}  & 2.6                        \\
& JCC~\cite{keuper2018motion}    &TPAMI &2018                   & 51.2 &54.5                      & -                        & 20.90\%                       & {37.00\%}                        & 25,937                        & 247,822                        & 1,802  & 1.8                    \\
        & SAS~\cite{maksai2018eliminating}    &CVPR &2019                   & 44.2         &{\color{blue} \textbf{57.2}}              & 76.4                      & 16.10\%                        & 44.30\%                        & 29,473                        & 283,611                        & 1,529  & {\color{blue} \textbf{4.8}}                        \\
      \hline
\end{tabular}
}
\vspace{-6pt}
\captionsetup{font=small}
\caption{\small \textbf{Quantitative results on MOT17.} The best scores of online and offline MOT methods are marked in {\color{red} \textbf{red}} and {\color{blue} \textbf{blue}}, respectively. }
\vspace{-2pt}
\label{tb:mot17}
\end{table*}

\begin{table*}
\centering\small
\setlength\tabcolsep{6pt}
\renewcommand\arraystretch{1.1}
\resizebox{0.99\textwidth}{!}{
\begin{tabular}{c|l||cccccccc}
\hline
\multirow{2}{*}{Aspect} &	\multirow{2}{*}{~~~~~~~~~~~~~~Module} &\multicolumn{8}{c}{validation set:~\{MOT17-09, MOT17-10\}} \\
\cline{3-10}
          &  & \multicolumn{1}{c}{MOTA$\uparrow$}   & \multicolumn{1}{c}{IDF1 $\uparrow$}   & \multicolumn{1}{c}{MT $\uparrow$}        & \multicolumn{1}{c}{ML $\downarrow$}        & \multicolumn{1}{c}{FP $\downarrow$}      & \multicolumn{1}{c}{FN $\downarrow$}       & \multicolumn{1}{c}{IDS $\downarrow$}
         \\ \hline
\multirow{2}*{Full model}&{\ourmodel~\textit{w.} TSA}   &\multirow{2}*{\textbf{53.0}} &\multirow{2}*{\textbf{61.9}} &\multirow{2}*{26.51\%} &\multirow{2}*{20.48\%} &\multirow{2}*{{969}} &\multirow{2}*{\textbf{7,236}} &\multirow{2}*{\textbf{56}}

\\
&Loss: $L_{\text{SOT}}$ (Eq.~\ref{eq:sot}) + $L_{\text{N-pair}}$ (Eq.~\ref{eq:npair_loss}) + $L_{\text{Iden}}$ (Eq.~\ref{eq:iden_loss})  &      &     &      &     &     &    &    &     \\
 \hline
\multirow{2}{*}{Loss}
&$L_{\text{SOT}}$  (Eq.~\ref{eq:sot}) + $L_{\text{Tri}}$ (Eq.~\ref{eq:tri_loss})
    &  51.7    &     50.9 &    24.10\%  &  \textbf{19.28\%}   &   922  & 7,483   &  87         \\
\multirow{2}{*}{(\textit{w/o.} TSA)}&$L_{\text{SOT}}$  (Eq.~\ref{eq:sot}) + $L_{\text{N-pair}}$ (Eq.~\ref{eq:npair_loss})
&  52.3    &     53.1 &    25.30\%  &  20.48\%   &   \textbf{851}  & 7,456   &  73      \\
&$L_{\text{SOT}}$  (Eq.~\ref{eq:sot}) + $L_{\text{N-pair}}$  (Eq.~\ref{eq:npair_loss}) + $L_{\text{Iden}}$ (Eq.~\ref{eq:iden_loss})
&  52.4    &     58.6 &    25.30\%  &  20.48\%   &   853  & 7,458   &  58      \\
 \hline
Structure &SOT \textit{only}  &  50.4    &     48.1 &  \textbf{27.71\%}  &  24.10\%   &   1,189  & {7,675}   &  138    \\
\hline
\end{tabular}
}
\vspace{-6pt}
\captionsetup{font=small}
\caption{\small \textbf{Ablation studies} on two validation sequences of MOT17. }
\vspace{-12pt}
\label{tb:ablation}
\end{table*}

\subsection{Performance on MOT Benchmark Datasets}
\label{sec:Performance}
\noindent\textbf{Quantitative and Qualitative Performance:} We evaluate our approach on the test sets of MOT16 and MOT17 benchmarks. The performance of our algorithm and other recent MOT algorithms are presented in Table \ref{tb:mot16} and Table \ref{tb:mot17}, where our lightweight~\ourmodel~model outperforms most online and even offline MOT algorithms according to the MOTA and IDF1 metrics. For instance, as shown in Table~\ref{tb:mot16}, we improve 1.7\% in MOTA, 5.6\% in IDF1 compared with KCF16~\cite{chu2019online}, which is an online algorithm taking KCF~\cite{henriques2014high} as the motion model.
Results from Table~\ref{tb:mot17} give another powerful support to the performance of our approach, on which we simultaneously achieve the better MOTA, MT, ML and FN against most published online and offline methods. In particular, FAMNet~\cite{chu2019famnet} is a recent work that also applies the Siamese SOT, and we surpass it in terms of both the MOTA and IDF1. Additionally, we improve Tracktor++~\cite{bergmann2019tracking} by 2.1\% according to the IDF1 metric, which validates the effectiveness of our unified model in dealing with occlusions and distractors. In a nutshell, our lightweight~\ourmodel~model achieves state-of-the-art performance, benefiting from the multi-task learning framework. Qualitative results of each sequence on the MOT17 test set are illustrated in Fig~\ref{fig:results}.

\noindent\textbf{Tracking Speed and Model Size:} {Our online MOT pipeline operates at a speed of around 5.0 fps on the test sequences of MOT17 with a 1080TI GPU, which is more efficient than most previous work, without losing the accuracy. Regarding the model size, for \cite{bergmann2019tracking} and \cite{zhu2018online}, it is 270M and 300M, respectively. In contrast, the whole size of our \ourmodel~is only around 30M, which is more suitable in source-constrained environments.}
\subsection{Ablation Studies}
\label{sec:ablation}
\vspace{-1pt}
In order to support the effectiveness of the proposed model, we conduct ablation studies on two sequences of the MOT17 training set, \ie,~\{MOT17-09, MOT17-10\}, and use other sequences for training.

{As shown in Table~\ref{tb:ablation}, compared with the basic SOT (last row), our full model achieves significantly better MOTA. This verifies the effectiveness of our whole framework (\S\ref{subsec:pipeline}). As our main motivation is to jointly learn the object motion prediction and affinity measure tasks, we next validate the ability of our model with multi-task learning (\textit{w/o.} the TSA module). We can observe that, compared with the framework trained only with the SOT, our method with the full loss (the next-to-last row) improves 13.8\%, 2.3\% and 59.4\% in terms of the IDF1, MOTA and IDS metrics, respectively. This demonstrates the significance of the integrated metric learning in addressing occlusions.
Third, we compare different combinations of losses within our \ourmodel~to validate the discriminative ability of the learnt features embedding. Considering results from the $2^{nd}$ to $4^{th}$ rows, we can find that jointly applying the N-pair loss and identification loss gives the best results according to the IDF1 and IDS metrics. Finally, we observe that the full model with the TSA module produces the best results, which further improves performance in terms of the MOTA, IDF1 and FN metrics. This indicates that the original shared features are less compatible with the tasks, {while the TSA module effectively stresses the task-aware context.}

\vspace{-2pt}
\section{Conclusions}
\label{sec:conclusions}
\vspace{-1pt}
This work proposed a novel online MOT model, named \ourmodel, aiming to perform the object motion prediction and affinity measure tasks in a unified network via multi-task learning. This is achieved by integrating the tracking loss and the metric learning losses into a triplet network during the training stage. The learnt feature not only enables the model to effectively measure the affinity in the data association phase, but also helps the SOT to distinguish the distractors during the tracklet generation phase. Compared with previous SOT based MOT approaches that train separate networks for the motion and affinity models, our method provides new insights by effectively improving the computation efficiency and simplifying the training procedure. Additionally, we extended our model with a TSA module to boost task-specific feature learning by emphasizing on different feature context. Extensive experimental results on the MOT16 and MOT17 benchmarks demonstrated the effectiveness of our lightweight model, which achieves competitive performance against many state-of-the-art approaches.

{\small\noindent\textbf{Acknowledgements}  This work was sponsored by Zhejiang Lab's Open Fund (No.~2019KD0AB04), Zhejiang Lab's International Talent$_{\!}$ Fund$_{\!}$ for$_{\!}$ Young$_{\!}$ Professionals, CCF-Tencent$_{\!}$ Open$_{\!}$ Fund and ARO grant W911NF-18-1-0296. 

{\small
\bibliographystyle{ieee_fullname}
\bibliography{motref}
}

\end{document}